\pdfoutput=1

\documentclass[11pt]{article}

\usepackage[final]{acl}

\usepackage{times}
\usepackage{latexsym}
\usepackage{booktabs}
\usepackage{url}
\usepackage[T1]{fontenc}

\usepackage[utf8]{inputenc}

\usepackage{microtype}

\usepackage{inconsolata}

\usepackage{graphicx}
\title{The Representational Alignment between Humans and Language Models is implicitly driven by a Concreteness Effect}

\author{Cosimo Iaia \\
  Goethe University Frankfurt  \\
  \And
Bhavin Choksi \\
  Goethe University Frankfurt \\
  \And
Emily Wiebers \\
  Goethe University Frankfurt \\
  \AND
Gemma Roig \\
 Goethe University Frankfurt \\
 Center for Brains, Minds and Machines, MIT \\
 Hessian.AI \\
  \And
Christian J. Fiebach \\
Goethe University Frankfurt \\
Brain Imaging Center \\
}

\begin{document}
\maketitle
\begin{abstract}

The nouns of our language refer to either concrete entities (like a table) or abstract concepts (like justice or love), and cognitive psychology has established that concreteness influences how words are processed. Accordingly, understanding how concreteness is represented in our mind and brain is a central question in psychology, neuroscience, and computational linguistics. While the advent of powerful language models has allowed for quantitative inquiries into the nature of semantic representations, it remains largely underexplored how they represent concreteness. Here, we used behavioral judgments to estimate semantic distances implicitly used by humans, for a set of carefully selected abstract and concrete nouns. Using Representational Similarity Analysis, we find that the implicit representational space of participants and the semantic representations of language models are significantly aligned. We also find that both representational spaces are implicitly aligned to an explicit representation of concreteness, which was obtained from our participants using an additional concreteness rating task. Importantly, using ablation experiments, we demonstrate that the human-to-model alignment is substantially driven by concreteness, but not by other important word characteristics established in psycholinguistics. These results indicate that humans and language models converge on the concreteness dimension, but not on other dimensions.

\end{abstract}

\section{Introduction}

\begin{figure*}[t!]

    \centering
    \includegraphics[width=0.9\linewidth]{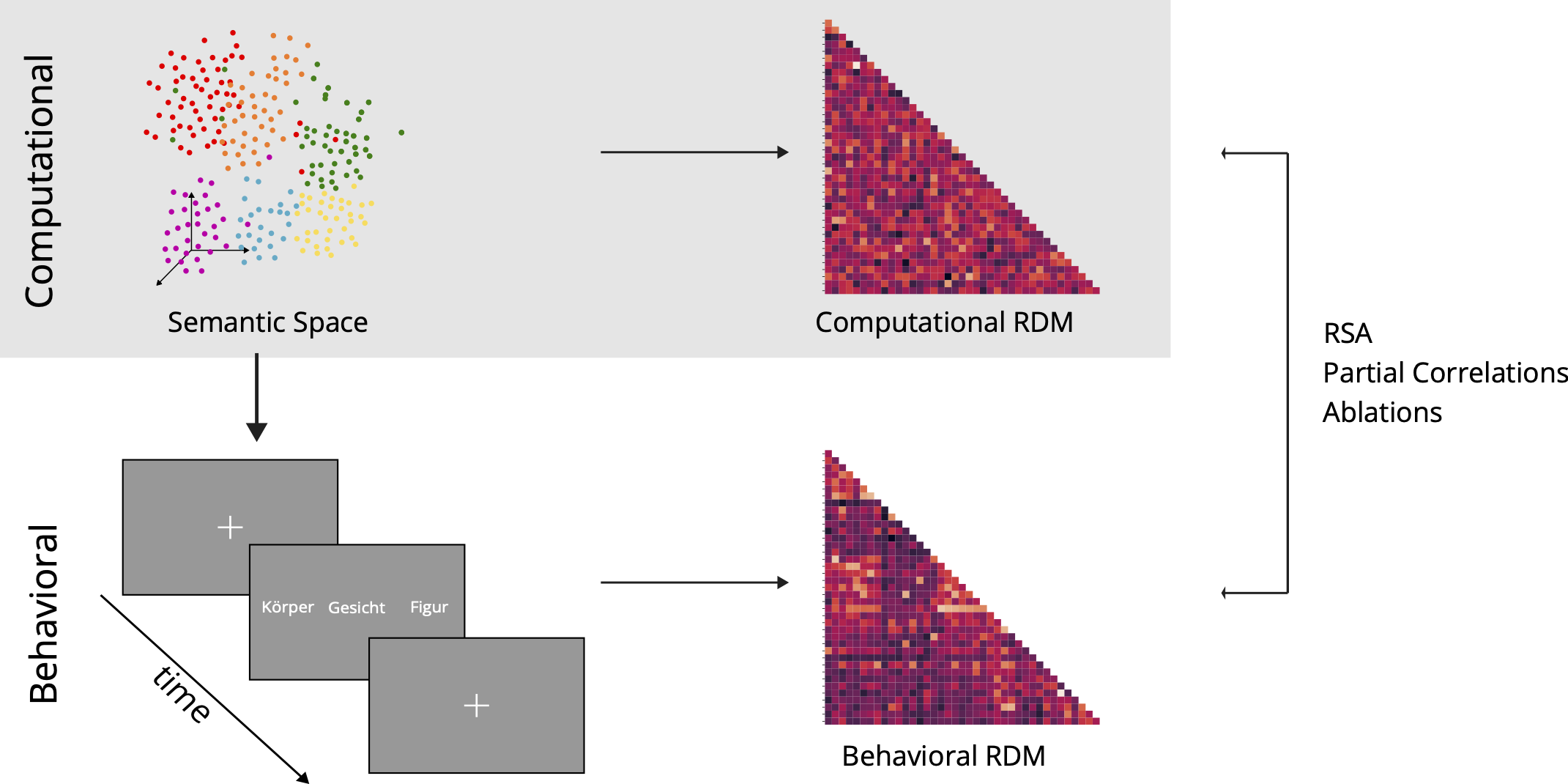}
    \caption{\textbf{Schematic of the approach}: Words are sampled from the semantic space (top left) and used for an odd-one-out task. The English translations of the German words are \textit{K\"orper} : Body, \textit{Gesicht} : Face, \textit{Figur} : Figure. 
    For a set of 40 words, we collected a total of 9880 odd-one-out choices. These were converted into a representational dissimilarity matrix (RDM) reflecting pair-wise semantic distances between words. For each language model, a similar computational RDM is created using the word embeddings for the 40 words. The two RDMs are then compared with each other.}
    \label{fig:enter-label}
\end{figure*}

The study of concreteness has a long history in Psychology and Neuroscience \cite{paivio1968concreteness}. Concreteness, usually measured through a rating task \citep[e.g.,][]{kanske2010leipzig, brysbaert2014concreteness}, refers to the extent to which a concept is related to sensory experience \cite{reilly2024we}.

Psycholinguistic research has established that whether words refer to concrete or abstract concepts influences word recognition behavior, indicating that concrete and abstract words may be represented differently \cite{solovyev2020concreteness}. Similarly, human neuroimaging studies have shown consistent differences in brain activation patterns elicited during the processing of concrete and abstract words \cite{fiebach2004processing, bucur2021ale}. 
Altogether, these and similar results strongly indicate that concreteness is an important dimension that critically influences how conceptual semantic knowledge is represented in the human mind and brain. 
The recent advancements in language models have also led to an increased interest in how they learn and represent concreteness. Without explicitly having learned about the concreteness dimension, language models can predict concreteness ratings with human-like degrees of accuracy \cite{koper2016automatically, martinez2025using, wartena2024estimating}. Language models can predict concreteness even cross-linguistically \cite{thompson2018automatic}, further supporting the universality of this property. However, the question of exactly how similarly language models and humans represent concreteness remains unanswered.

In this paper we address this question by measuring the alignment between the word representational spaces of humans and various language models. Using the method of Representational Similarity Analysis (see Section \ref{sec:rsa} below) we ask three fundamental questions: i) \textit{Is there a representational alignment between language models and humans in mental representations of single word meaning? ii) Do humans and language models implicitly represent concreteness? iii) Can concreteness independently explain the degree of agreement between the two representational spaces, i.e., is there a concreteness effect on the representational alignment?}

To address these questions, we first ran an odd-one-out task where we collected behavioral ratings for 9880 German word triplets to derive a representational space of word meanings and their similarities \cite{hebart2020revealing, turini2022hierarchical}. Participants (N = 40) were asked to determine which of the three words is least similar to the others (i.e., to determine the odd one). Given a triplet of items \textit{i, j, k}, the odd-one-out task allowed us to retrieve pairwise similarities between two items (for example, \textit{j,k}) compared to the same context item (\textit{i}), building an \textit{implicitly} derived representational space \cite{hebart2020revealing}. Importantly, word concreteness was neither explicitly probed nor in any other way relevant for this first behavioral task completed by the participants. However, words were selected such that they were semantically similar but varied in concreteness, as will be described in more detail in Section \ref{sec:methods}.
After the odd-one-out task, we asked participants to rate the same words used in the previous task for the abstract- vs. concreteness, to (i) validate the concreteness estimates used for designing the experiment, (ii) get a more accurate estimate of the explicit concreteness representation for our participants, and (iii) build an \textit{explicit} representational space for concreteness.

We then obtained the representations (or word embeddings) of the same words from popular language models and compared them to the behavioral data. We first show that there is a significant alignment between the representational space derived from human behavioral judgments and from the different language models. 
Additionally, both systems (human and language model representational spaces) proved to be independently aligned to a representational space derived from concreteness ratings, which can be considered a `gold standard' in psychology \citep[e.g.,][]{brysbaert2014concreteness}. 
Finally, to determine the extent to which concreteness drives the alignment, we first measured the partial correlations between the behavioral and model representations using concreteness as a control variable. 
Following \citet{oota-etal-2024-speech}, we also removed the maximum variance linearly explained by concreteness from the word embeddings. We then used these concreteness-ablated word embeddings to measure changes in the alignment with the behavioral ratings. 
Our results revealed that the alignment between the language models and humans is critically driven by concreteness. 
We publicly release our code and resources to replicate our findings.\footnote{Code will be available upon publication at \url{https://github.com/bhavinc/concreteness-in-LLMs}}

\section{Related work}
\label{sec:relatedworks}

The concreteness dimension has been shown to be relevant for the organization of word meaning and to influence semantic processing both behaviorally and neurally \cite{huang2015imaginative, montefinese2019semantic, bucur2021ale}. For example, concreteness affects memory performance \citep[concrete words are remembered better than abstract words,][]{fliessbach2006effect} and reaction times in lexical decision tasks \citep[concrete words are processed faster,][]{james1975role}. In a similar way, neuroimaging studies have shown different patterns of activation in the brain for abstract words in (mostly left-lateralized) and concrete words (bilaterally) \cite{binder2005distinct}. These, and related findings, are usually referred to as the concreteness effect \citep[cf.][]{bucur2021ale, solovyev2020concreteness, lohr2024does}. 

Given the importance of concreteness for word meaning processing, how language models represent this dimension has also been a major interest in the last years. A line of research in NLP has focused on the generation of automatic concreteness ratings from word embeddings \citep[e.g.,][]{koper2016automatically, wartena2024estimating} or by probing large language models to rate words on the concreteness scale \citep[e.g.,][]{martinez2025using}, with the primary objective to augment the ratings available within and among different languages. Thus, both humans and language models represent concreteness. Whether its representation is shared still lacks direct evidence.

Leveraging Representational Similarity Analysis\citep[RSA,][]{kriegeskorte2008representational}, \citet{bruera2023modeling} evaluated the impact of contextualized meaning of single words and phrases on neural processing of sentences, comparing the performance of GPT and a cognitive representational model based on concreteness. While providing evidence that GPT can represent concreteness to a certain degree, this study does not contribute to understanding whether its representation is shared between humans and language models for single word meaning.

In order to show to which degree the human-model alignment is dependent on concreteness, we combined an ablation approach to RSA: \citet{oota-etal-2024-speech} demonstrated that when linearly removing low level features (i.e., phonological features, number of characters) from language model representations, the brain-model alignment drops significantly in sensory cortices. Here, we applied a similar approach to show how removing concreteness from language models affects the representational alignment at the behavioral level. 

\raggedbottom

\section{Methods}
\label{sec:methods}

\subsection{Critical dimensions}
For our study we performed a behavioral experiment consisting of two tasks: an odd-one-out task and a concreteness rating task.
To inform our selection of word stimuli for both tasks, we used automatically generated concreteness ratings for German provided by \citet{koper2016automatically}. The aim of the rating task is to ensure that the representation of subjective concreteness will be as accurate as possible. Participants were asked to rate each word on a scale between 1 and 9 \cite{kanske2010leipzig}. To determine whether concreteness contributes unique variance in the representational alignment, we considered three more features commonly used in psychology, linguistics, and NLP related research: word frequency \cite{brysbaert2011word, brysbaert2018word}, word length (number of characters; as in \citet{oota-etal-2024-speech}), and orthographic similarity \citep[Orthographic Levishtein Distance 20, OLD20,][]{yarkoni2008moving}.
We used word frequency values on a log scale included in Subtlex-DE \cite{brysbaert2011word}, a corpus of words derived from German subtitles of movies. 
OLD20 was computed by taking the Levinstein distance of each word to each entry in Subtlex-De \cite{brysbaert2011word} and averaging the lowest 20 values (closest words).

\subsection{Selection of stimuli} 
\label{sec:selection stimuli}
Stimulus selection aimed at varying words along a concreteness/abstractness dimension, while controlling other semantic dimensions as much as possible. To ensure a broad representation of a plausible semantic space and the representation of the critical features, we selected 40 German nouns from a larger pool of nouns extracted from a book with the procedure that we describe in the next paragraph. Following \citet{pereira2018toward}, we performed spectral clustering of these nouns using the pretrained word embeddings from fastText: First, we constructed a cosine similarity matrix from the word vectors (fastText),  normalized between 0 and 1. We then zeroed out the diagonal of the matrix and normalized it row-wise to sum to 1. Following these steps, we performed a principal component analysis and applied k-means++ clustering (k = 19, empirically determined). 

To have a broad range in concreteness values, we chose the cluster with the highest variance for concreteness, and hand selected eight concrete words and eight abstract words (beyond 1 standard deviation from the mean for the cluster, mean = 4.95, std = 1.51) from the cluster while making sure that they were matched for word frequency, word length (between 5 and 9 characters), and OLD20 (all variables are within 1 std deviation from the mean of the cluster). Additionally, as a control condition, we selected eight frequent words, and eight infrequent words (beyond 1 standard deviation from the mean of the cluster, mean = 2.83, std = 0.77) while matching them for concreteness, length, and OLD20.  
Finally, we chose eight words that were closest to the centroid of the cluster, resulting in a total of 40 word items used for the experiment. As last step, all possible combinations of triplets were generated from the 40 nouns, resulting in 9,880 (= 40!/(3!*(40-3)!)) unique triplets \cite{turini2022hierarchical}.

\subsection{Experimental procedure}
Behavioral Data was collected at Goethe University in Frankfurt am Main, Germany. Participants (N = 40, age: M= 23.08, SD= 2.56, gender: F 27) were recruited through the internal system of the University, and were compensated either with money or credits. All participants were German native speakers and completed both tasks. The experimental procedure was approved by the Ethics Committee of the Goethe University (2019-17).
The number of participants was determined based on the optimal number of unique triplets to be presented, following \citet{turini2022hierarchical}'s experimental procedure. Word triplets were  presented to participants in randomized order. Each participants rated 247 triplets.

\subsection{Word embeddings}
For our experiments, we retrieved word vectors from popular distributional semantic models: We took 300-dimensional German word vectors provided in the \textit{fasttext} library \cite{grave2018learning}, trained on Common Crawl and Wikipedia using a Continuous Bag of Words approach (CBOW). We also used the German word2vec embeddings provided by \citet{mueller2015}. These embeddings were trained using skip-gram loss functions on a large corpus from the German Wikipedia ($\approx$ 651 Million words).

\begin{table*}
\centering
\begin{center}

\begin{tabular}{lrrrrrr}
\toprule
& Behavioral & Behavioral Conc & Automatic Conc & WFrequency & Length & OLD20 \\
\midrule
fastText & 0.51\textsuperscript{***} & 0.30\textsuperscript{***} & 0.24\textsuperscript{***} & 0.08\textsuperscript{*} & 0.00 & 0.06 \\
word2vec & 0.53\textsuperscript{***} & 0.28\textsuperscript{***} & 0.20\textsuperscript{***} & 0.09\textsuperscript{*} & 0.00 & -0.02 \\

BERT base & 0.24\textsuperscript{***} & 0.16\textsuperscript{***} & 0.07 & 0.17\textsuperscript{***} & 0.01 & 0.02 \\
BERT large & 0.37\textsuperscript{***} & 0.22\textsuperscript{***} & 0.20\textsuperscript{***} & 0.19\textsuperscript{***} & 0.02 & 0.14\textsuperscript{***} \\
GPT2 & 0.14\textsuperscript{***} & 0.06 & 0.01 & 0.00 & 0.08\textsuperscript{*} & -0.06 \\
\bottomrule

\end{tabular}
\caption{\textbf{Representational Alignment between the representations} of language models and the behavioral representations derived from the odd one out task (left column). Additionally, we show the correlations with the representational spaces based on concreteness (explicitly rated: Behavioral Conc; automatically generated: Automatic Conc) and word characteristics considered during stimulus construction (word frequency, word length, and orthographic similarity: OLD20). See Methods section for references. \textit{(*** p < 0.001, ** p < 0.01, * p < 0.05)}. Note that the behaviorally explicit concreteness ratings seem to perform better in representing concreteness than the automatically generated ratings.}

\label{tab:model-alignment}
\end{center}
\end{table*}

We also used the word embeddings learned by two popular large language models based on transformer architecture \cite{vaswani2017attention}: BERT and GPT2. BERT is optimized on a large corpus to predict the missing word within a context \cite{devlin-etal-2019-bert}. We used two german versions, BERT base and BERT large, provided by \citet{chan-etal-2020-germans}. 

GPT2, unlike BERT, is optimized using an autoregressive loss function to predict the next word given a set of previous context words \cite{radford2019language}. 
We used the german version of the model provided by MDZ Digital Library team at the Bavarian State Library \footnote{\url{https://huggingface.co/dbmdz/bert-base-german-uncased}}.

For both language models, we used the non-contextual word embeddings before any addition of positional embeddings or attentional blocks. Since the tokenization of words in these models is based on bytes, words are often broken down into smaller `tokens'. For our experiments, we average the embeddings across tokens to get the embeddings for individual words.

\subsection{Representational Similarity Analysis}
\label{sec:rsa}

We measured the alignment between the behavioral and language models' representational spaces using Representational similarity analysis (RSA) \cite{kriegeskorte2008representational}. RSA allows one to measure the alignment between two different representation spaces by first creating a Representational Dissimilarity Matrix (or RDM) using inter-pair distances. These RDMs characterize the representational geometry of each space using relative distances between points, and thus can be compared using some similarity metric (typically Pearson or Spearman correlation) \cite{kriegeskorte2008representational,kriegeskorte2013representational,sucholutsky2023getting}.

For the present purposes, we computed five different model RDMs from the word embeddings and language models described above, based on the pairwise cosine distance (1 - cosine similarity). A behavioral RDM reflecting the implicit similarity structure among words was derived from the participants' judgments in the odd one out task. To this end, following \citet{turini2022hierarchical}'s procedure, we assumed that when participants select a word in a triplet, the similarity between the two other words is high (coded with 1), and the similarity between the selected word and the other two is low (coded with 0). To retrieve the similarity of word pairs, we take the coded ratings (38 per combination) and average the choices.

Lastly, RDMs for further variables of interest (i.e., explicitly rated as well as automatically generated concreteness, word frequency, word length, OLD20; see above for further details) were constructed using euclidean distance, which in our case is the absolute difference between pairs of values. Spearman correlation is the used to measure the representational alignment.

\subsection{Removal of features from word vectors}
Following procedures described by \cite{oota-etal-2024-speech}, we investigated the impact of each variable of interest on the human-model alignment  by removing the respective variable from the word vectors derived from the language models. To remove the information (linearly) related to each feature from the word vectors we trained a Ridge regression model using \verb|sklearn| \cite{scikit-learn} to predict the word embeddings of a given language model (y) from the current variable of interests (X; e.g., concreteness). As a first step, we scaled both X and y by using the \verb|StandardScaler| implementation in \verb|sklearn|. For all models, except word2vec, we used 771 embeddings coming from all 5 models for our training set. As word2vec was missing pretrained embeddings for some words, we used 716 embeddings. (Note that there were also four word embeddings of the test set missing in word2vec). During training, we fine tuned the alpha parameter for the Ridge regression based on R\textsuperscript{2}. To remove concreteness from word embeddings, we used the automatically generated ratings provided by \citet{koper2016automatically} for both train and test sets.

As discussed by \citet{oota-etal-2024-speech}, this approach to removing features from embeddings assumes a linear relationship between the feature variables and the embedding. 

We computed word-specific residuals for the 40 test items by subtracting predicted word vectors from the original, model-derived word vectors, resulting in a new word embeddings in which the information related to the respective feature is not present any more. We then used the residual word vectors to construct new RDMs (ablated RDMs) and re-ran the RSA, now between ablated model RDMs and the behavioral RDM derived from the odd-one-out task. This feature removal (or ablation) was repeated for each language model, and the entire procedure was independently conducted for all features of interest, i.e., concreteness, word frequency, word length, and orthographic similarity (OLD20). The effect of feature removal on model alignment was assessed statistically by comparing the resulting RSA correlation against the `base correlation', i.e., the RSA correlation obtained for the full model RDM, using the Williams Test, a test for comparing two correlations that have one shared variable (in this analysis, the behavioral RDM is shared between `base' and ablated correlations) \cite{williams1959comparison, diedenhofen2015cocor}.

\section{Results}

\subsection{Human-model alignment}

\begin{figure*}[th!]
    \centering
    \includegraphics[width=0.9\linewidth]{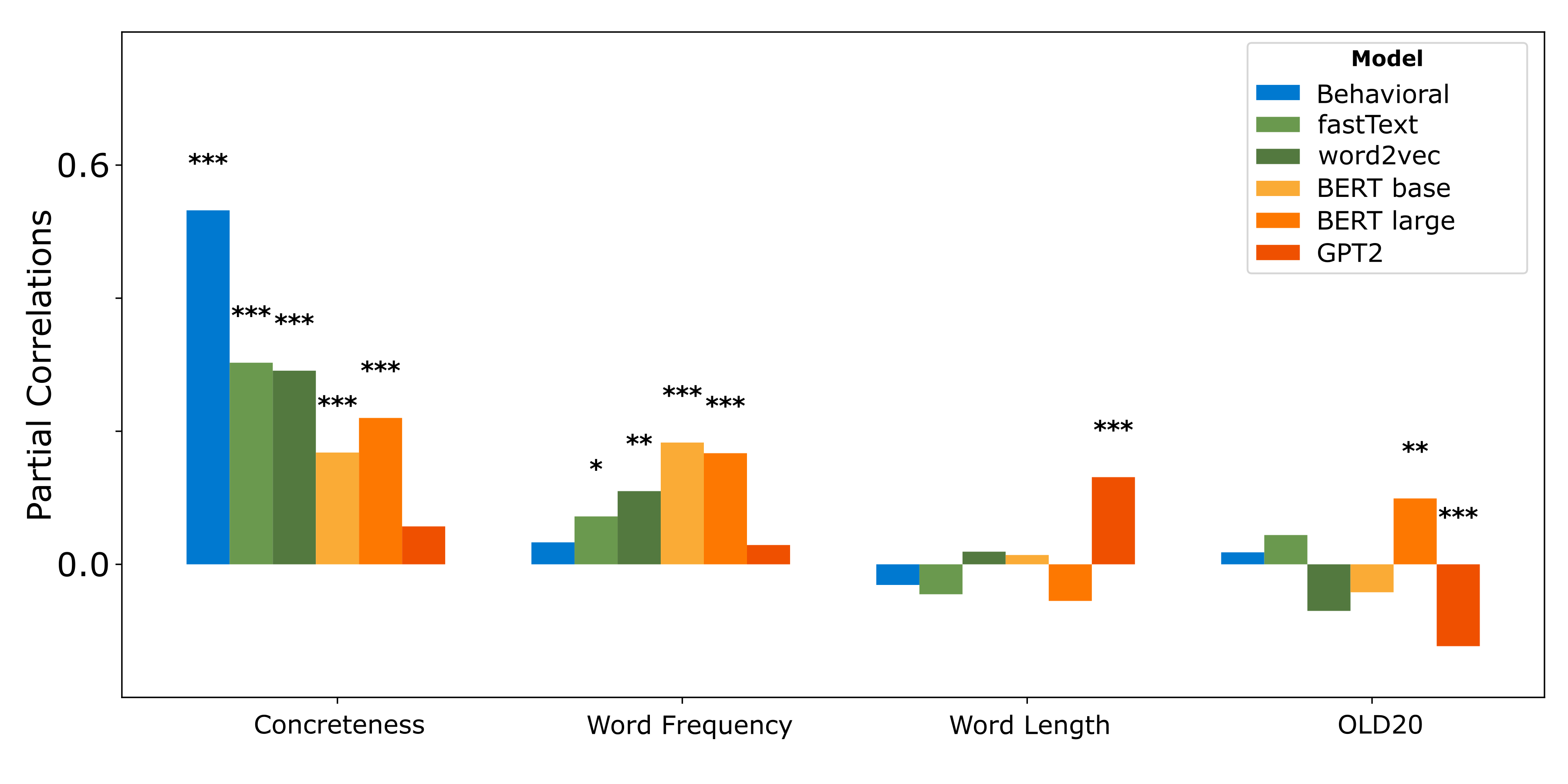}
    \caption{\textbf{Partial correlations for the behavioral model (odd-one-out) and the computational models (language models):} The representational space derived from the odd-on-out (in blue) is only correlated to the rated concreteness space, while language models (other colors) are aligned to other feature spaces as well. The representational spaces derived from all language models but GPT2 (in red) show alignment not only to concreteness but also to word frequency. GPT2, instead, is correlated to word length and OLD20. \textit{(*** p < .001, ** p < .01, * p < .05)}.}
    \label{fig:partial_correlations}
\end{figure*}

Given that stimuli were selected based on automatically generated concreteness ratings, we first examined the validity of this approach by comparing them with actual concreteness ratings obtained from our participants.  Concreteness ratings were averaged across participants, separately for each word. The correlation analysis revealed a very high agreement between automatic and subjective ratings (Pearson r = 0.86), in a comparable range to the original report of the automatically generated ratings \citep[r = .825; cf.][]{koper2016automatically}.

Our main analyses involved RSAs  between the computational RDMs derived from language models and the representational spaces based on (i) the implicit similarity estimates derived from behavioral odd one out ratings, and (ii) explicit features , i.e., concreteness, word frequency, length, and orthographic similarity. We observe significant alignment between computational language models and implicitly derived human similarity representations, with the highest correlation values for the static distributional models (fastText, word2vec; both $\rho$ > .5) and lower, albeit significant, correlations for the contextualized models (BERT base, BERT large, GPT2) as shown in Table \ref{tab:model-alignment}. Specifically, GPT2 showed the worst alignment to the implicit representational space ($\rho$ = .14) and BERT models showed alignment correlations of $\rho$ = .24 and $\rho$ = .37.

Even though our main focus is on the implicit similarity-based representational space derived from the participants' odd one out ratings, we also assessed the alignment between language models and explicit feature ratings. As Table \ref{tab:model-alignment} shows, there is significant alignment between model representations and a representational geometry purely derived from word concreteness, either explicitly rated (2nd column) or automatically generated (3rd column). Interestingly, RSA correlations are higher for behavioral than for automatic concreteness ratings, but both are substantially smaller than the similarity-based alignment correlations. Word frequency shows even lower degrees of alignment, where BERT-based embeddings show best alignment. Interestingly, GPT2 word embeddings show no alignment with either concreteness or word frequency. Lastly, the `lowest level' word features - their length and orthographic similarity - do not seem to be represented in the majority of language models.

\subsection{Alignment with Feature-based Representations}

We next aimed to explore whether or not word concreteness, our primary variable of interest, is represented in the behaviorally derived similarity-based representational geometry and in the language models, and whether or not these representations are independent of other fundamental word features. To this end, we ran partial correlations to assess the degree of alignment (i) between the (implicit), task-derived behavioral similarity representations and further word features considered in this study (word concreteness, frequency, length, and orthographic similarity, as well as (ii) between the computational representations derived from language models and these word feature representations. Partial correlations were computed using the Python package \verb|pingouin| \cite{vallat2018pingouin}. These analyses showed a significant correlation of $\rho$ = .53 (p < .001) between the  behavioral RDM from the odd-one-out task and the RDM representing subjective concreteness ratings while controlling for word frequency, length, and OLD20 (see Fig. \ref{fig:partial_correlations}, left-most blue bar). The implicit behavioral representations are not correlated with any other feature RDM if concreteness is partialled out (as shown by the other blue bars in Fig. \ref{fig:partial_correlations}), suggesting that the control variables had no impact on the representational alignment. For the computational RDMs, all models but GPT2 showed unique variance explained by concreteness (fastText $\rho$ = .30, word2vec, $\rho$ = .29, BERT base $\rho$ = .17, BERT large $\rho$ = .22, GPT2 $\rho$ = 0.06), when controlling for word frequency, length, and OLD20. Compared to \citet{bruera2023modeling}, we did not find a correlation between the single word concreteness representational space and GPT2. Note that this might be due to several reasons, such as a different procedure of the extraction of representations (i.e., averaging across different layers and using sentences as input rather than single words as in our study), or the training material of the model (GPT2 trained for Italian vs German in our study). These results provide evidence that both the implicit behavioral representation and the computational representation of single word meaning for most models is aligned to an explicit concreteness representational space.

\begin{figure*}[th!]
    \centering
    \includegraphics[width=0.9\linewidth]{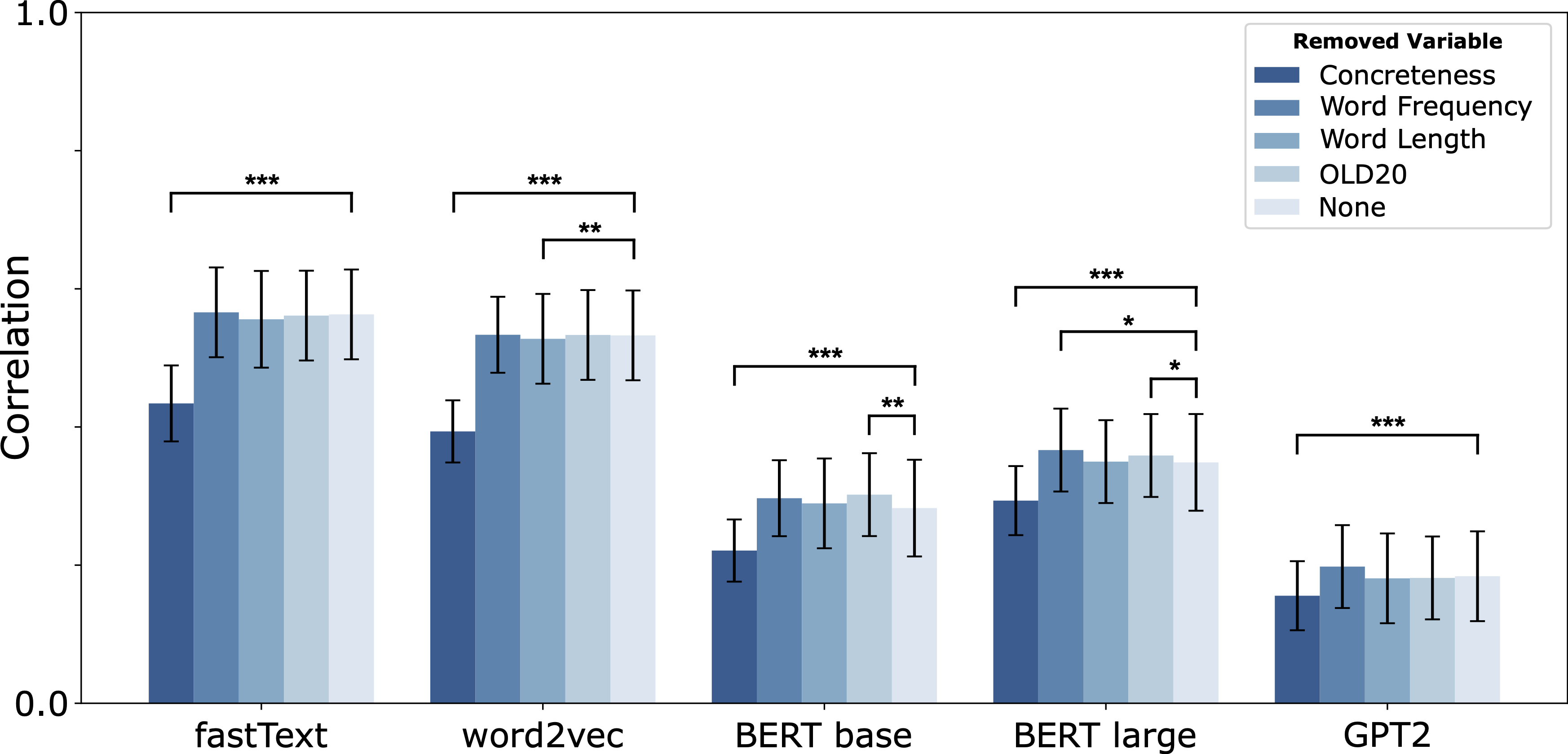}
    \caption{\textbf{Representational Similarity Analysis after removing each feature:} Compared to the original correlations between the non-ablated computational representation (lightest shade of blue)  and the representation derived from the odd-one-out task, the biggest drop is observed when removing concreteness (dark blue) for all language models. \textit{(Williams' test, *** p < .001, ** p < .01, * p < .05)}}
    \label{fig:ablated_correlations}

\end{figure*}

\subsection{Impact of feature removal on human-model alignment}

The main analysis of this study investigates the importance of concreteness--and potential other word features--for the alignment between human-derived and model-derived geometries of word representations. To this purpose, we implemented an ablation approach that removed individual word features from behavioral and model RDMs (as described in detail in the Methods section). Representational similarity was re-analyzed after removing each feature from the representational spaces of the different language models. The biggest drop compared to the above-reported alignment with the full models was observed only when removing concreteness, regardless of the architecture of the model (see Fig. \ref{fig:ablated_correlations}). This effect was consistent and significant (p < .001) for all the language models we analyzed. On average, we observed a 20.6 \verb|%| drop from the initial correlation values, with the biggest drop for word2vec (26 \verb|%|). In addition, removing word length significantly affected the alignment for word2vec, removing word frequency affected BERT large, and removing OLD20 affected alignment for both BERT base and BERT large. These changes in model alignment were significant (p < .05) but all below 7.6 \verb|%| in magnitude. To summarize, our results indicate that representational alignment is greatly affected by concreteness but not by any other tested variable. To further strengthen our results, we ran a control analysis comparing our results for concreteness with a set of other semantic variables, i.e., imageability, valence, and arousal, provided once again by \citet{koper2016automatically}, showing that none of them resulted in a decreased alignment as much as for concreteness (see Appendix \ref{sec:appendix}).

\section{Discussion and Conclusion}
By investigating the representational alignment between humans and language models, we here show that concreteness plays an important role for the internal representations of both systems--without being explicitly trained (models) or probed (humans) on concreteness--and that concreteness contributes in a critical manner to their alignment. This conclusion is supported by the observations 1) that both systems are independently aligned to a representational space based on explicit concreteness ratings, 2) that removing the concreteness feature from the semantic spaces of the language models decreases their alignment with the human data, and 3) that removal of `lower-level' orthographic or lexical features has no comparable influence on the human-to-model alignment. Taken together these results show a concreteness effect in the representational alignment between human word representations and language models. 

These results lead to the conclusion that the representation of concreteness converges between humans and machines. 

Previous work did not directly address the question whether humans and language models have a shared representation of concreteness \citep[cf.][]{bruera2023modeling}.

By showing that a substantial portion of the representational alignment is explained by concreteness, we provide evidence that the representation of concreteness is indeed shared between humans and language models. This conclusion is strengthened by the fact that concreteness emerged implicitly in both representational spaces, i.e., without models being explicitly trained to represent concreteness, and in a human behavioral similarity space derived from a task that did not focus participants' attention on word concreteness. 

The importance of concreteness for alignment is further strengthened by our finding that other features that strongly influence word recognition behavior (like word frequency, orthographic similarity, and word length) are represented differently between humans and machines. While \citet{oota-etal-2024-speech} showed that for spoken language processing the removal of low-level speech features has an effect on the brain-model alignment in sensory cortices, we found that such `lower-level' features had no impact on the representational alignment at the behavioral level. One reason for this might be a more limited variation of the low-level features (such as word length) in our experimental design (see Section \ref{sec:selection stimuli}), given the work with a limited stimulus set as opposite to Oota \textit{et al.}’s approach using a larger, naturalistic stimulus set. Thus, one important extension of the present approach would be to replicate our results in larger datasets. It is important to highlight that non-semantic features can be represented differently even between language models: for example, \citet{lenci2022comparative} reports lower correlation values for the alignment between different language models when comparing very frequent words to mid or low frequency words, also highlighting difference in model-to-model agreement when looking at different Parts-of-Speech. These findings highlight that, while language models are able to predict human behavior, and neural representations of word meaning, they differ in the representation on some dimensions, suggesting that caution is needed when using a distributional model as a model of human semantic processing.

While we show that concreteness is a relevant dimension organizing semantic representations in the human mind, also implicitly, and that language models represent concreteness in highly similar ways, the question arises how the abstract-concrete dimension is organized in the human brain. Thus, extending the present approach to functional neuroimaging data and model-to-brain alignment will be highly fruitful to improve our understanding of the organization of semantic representations in the human brain. Also, whether the representational differences between abstract and concrete words--and the concepts they refer to--may themselves be driven by other factors, is an important question emerging from the current results. For example, \citet{lohr2024does} suggested that word imageability and the availability of contextual information related to words may drive concreteness effects (but see Appendix \ref{sec:appendix}). Thus, our work may open up important avenues for exploring the nature of the concreteness dimension in human semantic representations in more depth. To this end, our approach might be fruitfully combined with established strategies of identifying fine-grained, meaningful semantic dimensions \citep[e.g.,][]{binder2016toward, hebart2020revealing}.

\newpage
\section{Limitations}

While the words in our stimulus set were carefully selected to be a representative sample from a larger pool of items, the current work remains limited in the total number of words chosen for the behavioral study. This was due to practical constraints of the word triplet task, as the combination of all words into an exhaustive set of triplets results in very large numbers of behavioral ratings required. We compensated for this by the very controlled selection of stimuli. Similarly, due to practical constraints, the maximum range of the word frequency values used does not perfectly match the natural range of word frequency in a larger corpus, and low frequency words are more strongly represented in the stimulus set than high frequency words. Thus, though our results are supported by large evidence from both Psychology and NLP, one goal for future research could be replication studies with different sets of words (including in other languages) to ensure generalizability of our results.

More importantly, the current study aims to characterize the alignment between the two spaces in terms of concreteness.
However, rather than considering this a limitation, we suggest that the present work opens up an interesting novel way of investigating in further depth the nature of concreteness effects. Similarly, the current definition of concreteness rests on individual non-contextualized words. How would the concreteness of words evolve within a context, and its effect on the alignment with the language models remains a question still not investigated enough \citep[cf.][]{bruera2023modeling}. While our results do not directly address this debate, we hope the provision of our data and further analysis into it will aid future research in addressing these questions.

\section*{Acknowledgments}
The authors gratefully acknowledge the funding support of the Deutsche Forschungsgemeinschaft
(DFG) - DFG Research Unit FOR 5368 (project number 459426179) for GR (DFG RO 6458/2-1) and CJF (DFG FI 848/9-1).
\bibliography{custom}

\clearpage
\newpage
\appendix

\section{Appendix}
\label{sec:appendix}

To further characterize the nature of the representational alignment, we also used imageability, valence, and arousal in combination with the ablation approach. The ratings were provided by \citet{koper2016automatically}. Note that Imageability is known to be highly correlated to concreteness. For our set of 40 words, we report a Pearson correlation of .93 between imageability and concreteness. Interestingly, we observe a consistent drop only for concreteness but not for other features. We found, however, a significant change of imageability for all language models, except word2vec. For both fastText, and GPT2, valence is significantly different than the base correlation as well. Taken together, these results indicate that concreteness still plays a more critical role in the human-model alignment.

\begin{figure*}[th!]
    \centering
    \includegraphics[width=1\linewidth]{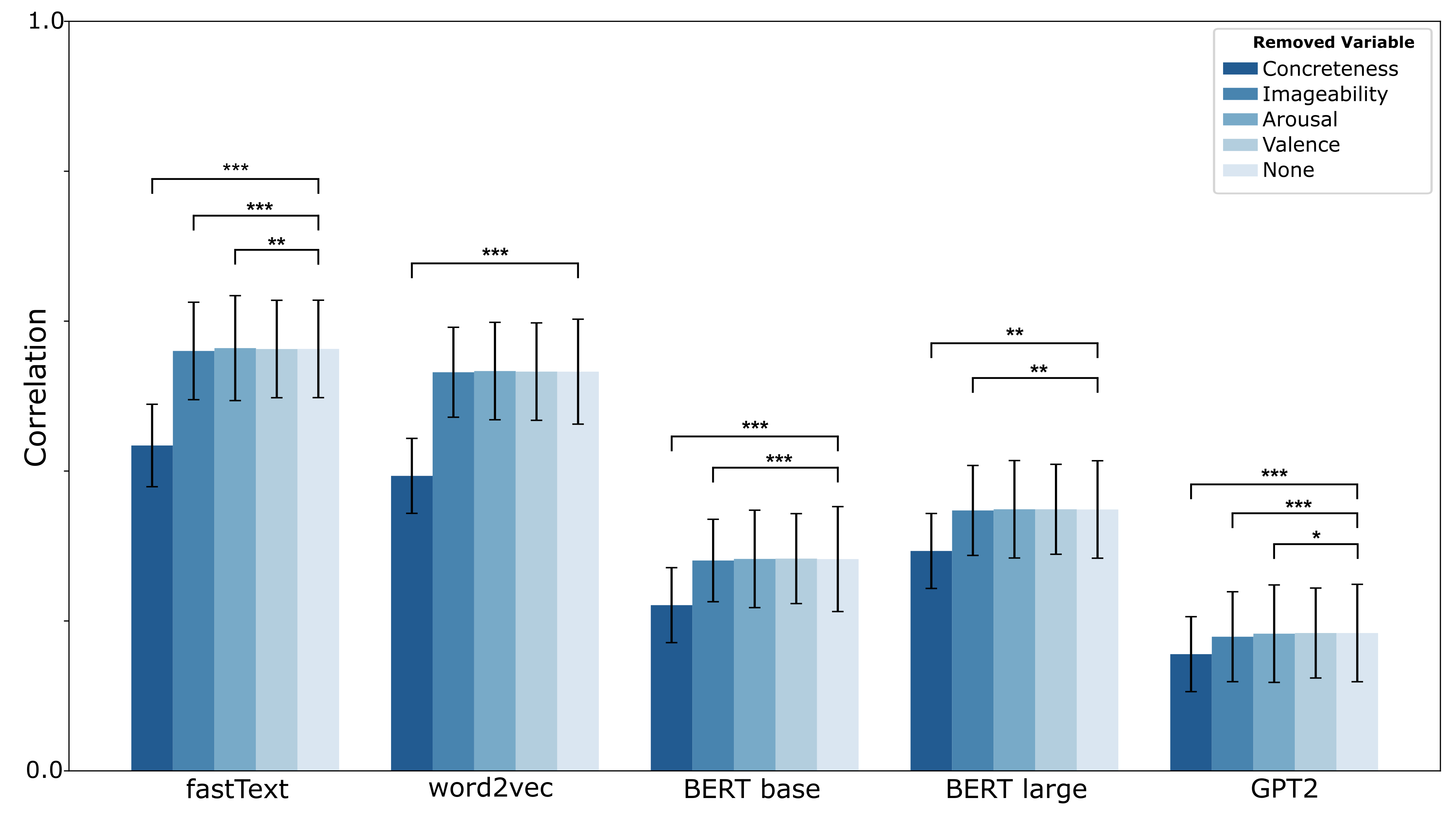}
    \caption{\textbf{Control Analysis: Representational Similarity analysis after removing further semantic features}. The ablation approach reported in the main paper was repeated with further semantic dimensions, i.e., word imageability, word arousal, and word valence. Compared to the base correlations, removing concreteness resulted in the biggest drop for all language models.}
    \label{fig:ablation_imageability}
\end{figure*}


\end{document}